%% file: main.tex
\documentclass[10pt,twocolumn,letterpaper]{article}

\usepackage[pagenumbers]{iccv} 

\input{preamble}

\definecolor{iccvblue}{rgb}{0.21,0.49,0.74}
\usepackage[pagebackref,breaklinks,colorlinks,allcolors=iccvblue]{hyperref}

\def\paperID{4312} 
\def\confName{ICCV}
\def\confYear{2025}

\title{\ourstitle}

\author{
Seongchan Kim$^{1*}$ \qquad Woojeong Jin$^{1*}$ \qquad Sangbeom Lim$^{2*}$ \qquad
Heeji Yoon$^{1*}$ \\ Hyunwook Choi$^{2}$ \qquad Seungryong Kim$^{1\dagger}$ \\[5pt]
 \qquad $^{1}$KAIST \qquad $^{2}$Korea University
}

\usepackage{colortbl}
\usepackage{xcolor}
\usepackage{multirow}
\usepackage{subcaption}
\usepackage{amssymb}
\usepackage{pifont}
\usepackage{blindtext}
\usepackage{array}
\newcolumntype{L}[1]{>{\raggedright\let\newline\\\arraybackslash\hspace{0pt}}m{#1}}
\newcolumntype{C}[1]{>{\centering\let\newline\\\arraybackslash\hspace{0pt}}m{#1}}
\newcolumntype{R}[1]{>{\raggedleft\let\newline\\\arraybackslash\hspace{0pt}}m{#1}}
\usepackage{booktabs}
\usepackage{graphicx}
\usepackage{relsize}
\usepackage{lipsum}

\begin{document}
\maketitle

\input{sec/0_abstract}
\input{sec/1_introduction}

\input{sec/2_related_work}
\input{sec/3_method}
\input{sec/4_experiments}
\input{sec/5_conclusion}
{
    \small
    \bibliographystyle{ieeenat_fullname}
    \bibliography{main}
}
\input{supplementary}

\end{document}

%% file: preamble.tex
\newcommand{\ours}{SOLA\xspace}
\newcommand{\oursfullname}{\textbf{S}election by \textbf{O}bject \textbf{L}anguage \textbf{A}lignment}
\newcommand{\ourstitle}{Referring Video Object Segmentation via Language-aligned Track Selection}

\newcommand{\cmark}{\ding{51}}
\newcommand{\xmark}{\ding{55}}

\newcommand{\paragrapht}[1]{\noindent\textbf{#1}}

\definecolor{changecolor1}{RGB}{255,0,0}
\definecolor{lightblue}{rgb}{0.9, 0.95, 1.0}
\definecolor{lightgray}{rgb}{0.95, 0.95, 0.95}

%% file: sec/0_abstract.tex
\begin{abstract}
Referring video object segmentation (RVOS) requires tracking and segmenting an object throughout a video according to a given natural language expression, demanding both complex motion understanding and the alignment of visual representations with language descriptions. Given these challenges, the recently proposed Segment Anything Model 2 (SAM2) emerges as a potential candidate due to its ability to generate coherent segmentation mask tracks across video frames, and provide an inherent spatio-temporal objectness in its object token representations.
In this paper, we introduce \textbf{\ours} (\oursfullname), a novel framework that leverages SAM2 object tokens as compact video-level object representations, which are aligned with language features through a lightweight track selection module. To effectively facilitate this alignment, we propose an IoU-based pseudo-labeling strategy, which bridges the modality gap between SAM2 representations with language features.
Extensive experiments show that \ours achieves state-of-the-art performance on the MeViS dataset and demonstrate that \ours offers an effective solution for RVOS.
Our project page is available at: \href{https://cvlab-kaist.github.io/SOLA}{\texttt{https://github.com/cvlab-kaist/SOLA}}.
\end{abstract}

%% file: sec/1_introduction.tex
\section{Introduction}

\input{fig/main_teaser}

Referring video object segmentation (RVOS)~\cite{ding2023mevis, gavrilyuk2018actor, khoreva2019video, seo2020urvos} aims to identify and segment a specific object throughout a video sequence based on a natural language expression. RVOS has recently attracted significant research interest due to its broad applicability in various fields, including interactive video editing and human-robot interaction. However, RVOS presents several challenges, as the model must integrate both natural language comprehension and visual understanding at both the scene and object levels.

Recently, segment anything model (SAM)~\cite{kirillov2023segment} has emerged as a powerful models in the field of segmentation, demonstrating remarkable performance across various tasks. In particular, SAM2~\cite{ravi2024sam2} excels in generating segmentation masks across video frames, an essential capability for accurate object tracking and motion modeling in dynamic environments. Since generation of segmentation mask tracks inherently requires maintaining object identity over time, SAM2 implicitly captures temporal-aware object information at the video level. This characteristic makes SAM2 highly relevant for addressing the challenges of RVOS.
Despite this potential, certain challenges arise in directly applying SAM2 to RVOS, as SAM2 is designed solely for segmentation and lacks an understanding of natural language. Moreover, while SAM2 effectively encodes object information, how to exploit this inherent knowledge remains an important question.

Based on the observation that SAM2’s object token representations inherently capture temporally consistent object regions, our \textbf{\ours} (\oursfullname) framework leverages these tokens as compact video-level representations, enabling robust motion understanding—crucial for RVOS. To associate objects with language, we introduce a lightweight \textit{language-aligned track selection module}, effectively bridges the modality gap between SAM2’s object token representations and language features, as illustrated in Figure~\ref{fig:main_teaser}. Notably, \ours utilizes solely precomputed object tokens, enabling efficient training on a single GPU while preserving SAM2’s robustness and generalizability. Additionally, we introduce a novel training strategy that leverages IoU (Intersection over Union)-based pseudo-labels to supervise a simple binary classification objective, complemented by a contrastive loss designed to highlight detailed motion patterns. To utilize precomputed object tokens during training, we generate pseudo-labels based on the IoU between the mask track corresponding to the precomputed object tokens and the ground truth mask track associated with the given referring expression.

In our experiments, we evaluate our method on standard RVOS benchmarks, including MeViS~\cite{ding2023mevis}, Ref-YouTube-VOS~\cite{li2018referring}, and Ref-DAVIS~\cite{KhoRohrSch_ACCV2018}. Our framework achieves state-of-the-art performance on MeViS, demonstrating its effectiveness in tracking object sequences guided by complex language expressions. Additionally, our method exhibits strong generalizability and robustness across diverse settings, including zero-shot and combined dataset evaluation. Our method achieves both high-quality tracking and effective multi-modal alignment, excelling in both quantitative and qualitative evaluations.

Our main contributions are as follows:
\begin{itemize}
    \item We propose \ours, a novel framework that, for the first time, utilizes SAM2’s object token representations for RVOS. We hypothesize that these tokens inherently encode temporal-aware objectness, enabling effective motion modeling of an object.
    \item By introducing a lightweight language-aligned track selection module that relies exclusively on object tokens. This approach allows for the use of precomputed tokens, enabling efficient training on a single GPU.
    \item We adopt a novel training strategy that utilizes IoU-based pseudo-labels for object tokens, enabling our language-aligned track selection module to effectively bridge the modality gap between SAM2’s object representations and language features.
    \item Our method achieves new state-of-the-art results on the MeViS dataset~\cite{ding2023mevis} and demonstrates strong generalization on Ref-YouTube-VOS~\cite{li2018referring} and Ref-DAVIS~\cite{KhoRohrSch_ACCV2018}, excelling in both quantitative and qualitative evaluations.
\end{itemize}

\input{fig/main_figure}

%% file: fig/main_teaser.tex
\begin{figure}[t]
    \centering
    \includegraphics[width=\linewidth]{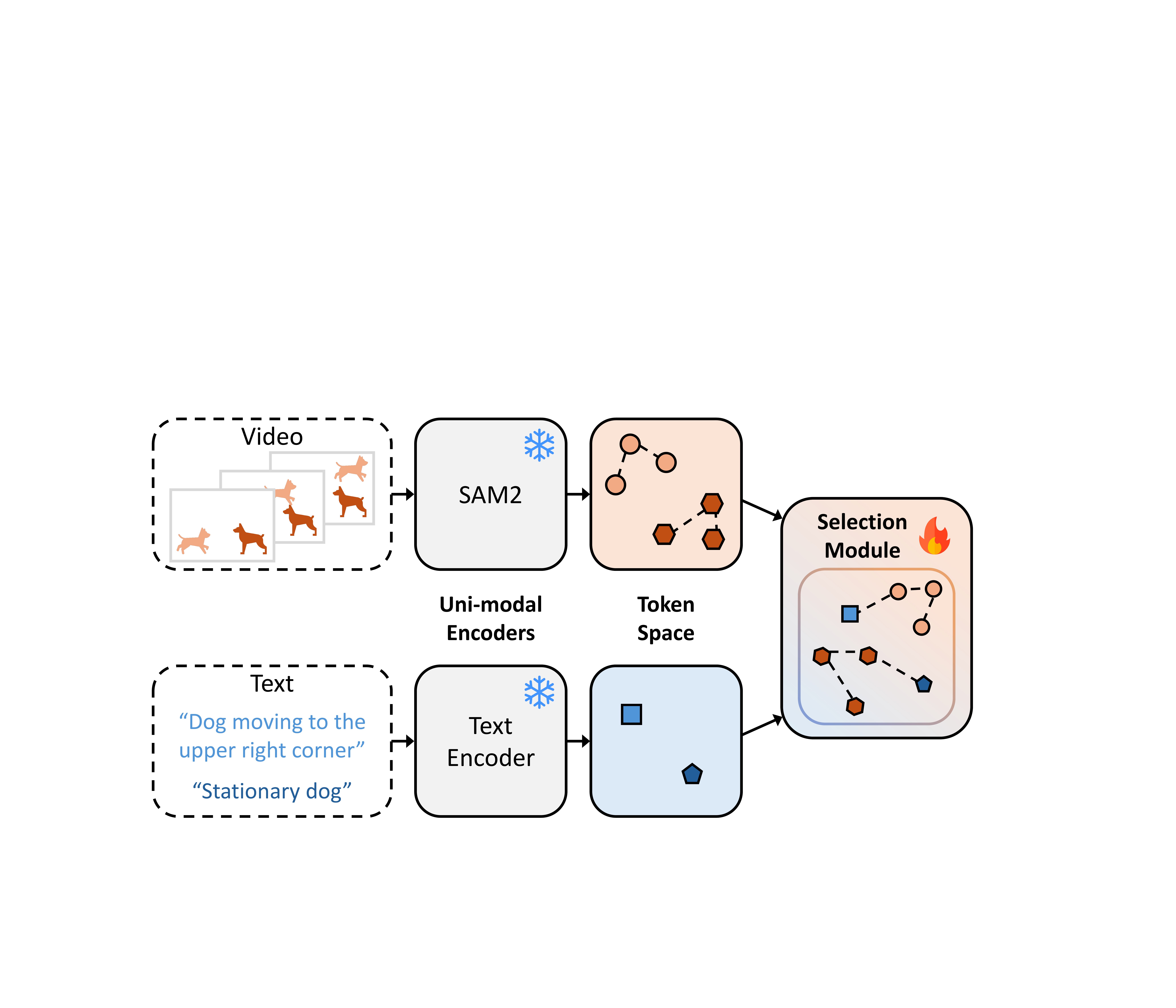}
    \vspace{-10pt}
    \caption{\textbf{Teaser.} Our method effectively bridges the modality gap by aligning the features obtained from fully frozen uni-modal encoders: the video segmentation model such as SAM2~\cite{ravi2024sam2} and the text encoder such as RoBERTa~\cite{DBLP:journals/corr/abs-1907-11692}. By directly leveraging the token representations, our approach achieves lightweight multi-modal alignment while significantly reducing the number of trainable parameters.}
    \label{fig:main_teaser}
    \vspace{-10pt}
\end{figure}

%% file: fig/main_figure.tex
\begin{figure*}[ht]
    \centering
    \includegraphics[width=\linewidth]{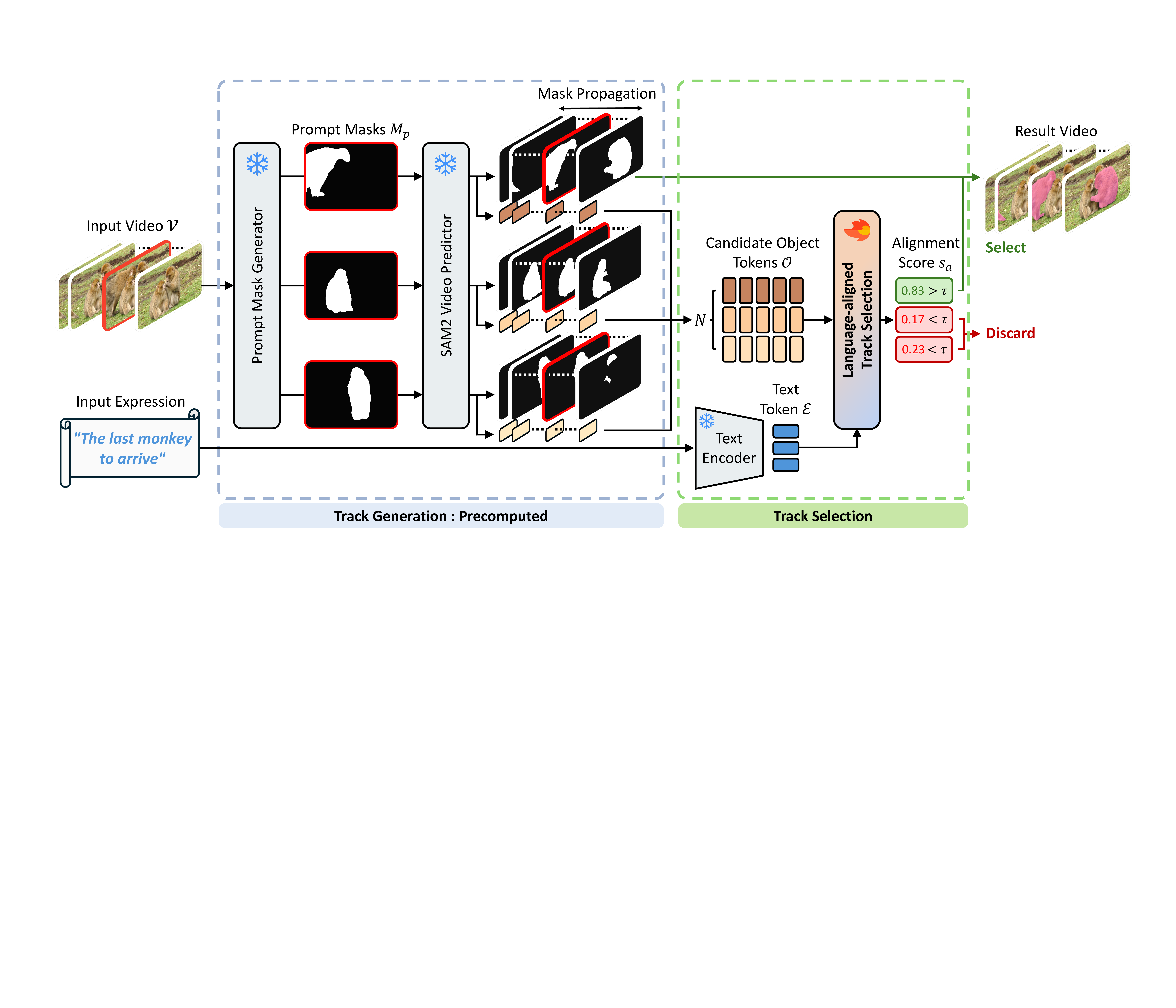}
    \caption{\textbf{Overall pipeline of the proposed \ours framework.} Our method selects the correct object mask track among candidates via a language-aligned track selection module. We first generate candidate mask tracks and corresponding object tokens from the fully frozen SAM2. These tokens are then aligned with language expressions, producing alignment scores that indicate selection probabilities. Mask tracks with scores above a predefined threshold are selected and merged into the final binary segmentation mask. By leveraging precomputed object tokens from SAM2, our approach minimizes trainable parameters, enabling efficient training on a single GPU.}
    \label{fig:main_figure}
    \vspace{-10pt}
\end{figure*}

%% file: sec/2_related_work.tex
\section{Related Work}

\paragraph{Referring video object segmentation.}
RVOS requires segmenting objects by capturing both action and appearance from video sequences based on a given expression. RVOS was first introduced by Gavrilyuk et al.~\cite{gavrilyuk2018actor} with the A2D-Sentences benchmark. Since then, RVOS has garnered significant attention, leading to the development of benchmarks such as Ref-YouTube-VOS~\cite{li2018referring}, Ref-DAVIS~\cite{KhoRohrSch_ACCV2018}, and MeViS~\cite{ding2023mevis}.

Recently, query-based models~\cite{wu2022language, botach2022end, ding2023mevis, he2024decoupling, miao2024htr} have achieved impressive performance by leveraging object query tokens. These tokens are expected to capture spatial properties, appearance, and temporal dynamics while maintaining temporally consistent object mask tracks. Other approaches~\cite{han2023html, li2023robust} enhance language alignment by employing object tokens pre-aligned with language features.
Thereby, solving RVOS demands a model that ensures temporal consistency while effectively linking textual descriptions with visual representations containing various object information.

\paragrapht{Segment anything model.} 
SAM~\cite{kirillov2023segment} is known as a breakthrough in foundation models for image segmentation, with a unique ability to segment any object within an image using interactive prompts. Building on SAM, SAM2~\cite{ravi2024sam2} extends its capabilities to video segmentation through a memory-based transformer. SAM2’s memory stores information about target objects and past interactions, enabling it to perform segmentation more accurately and efficiently while maintaining strong generalization performance.

There are previous approaches~\cite{li2023refsam, huang2024unleashing} that utilizes SAM or SAM2 in RVOS task. However, these approaches predominantly use them only at the prompting level, treating them merely as powerful mask generation tools without tapping into their rich internal representations for more advanced video-level object understanding.
Ref-SAM~\cite{li2023refsam} processes textual inputs by projecting them into sparse and dense prompts, but these prompts mainly tied to image-level, propagating from a selected object through an implicit tracking module. As a result, it struggles to handle complex motion across an entire video sequence or to differentiate among objects of similar classes.
Similarly, AL-RefSAM 2~\cite{huang2024unleashing} assigns the spatio-temporal reasoning capability on GPT-4~\cite{achiam2023gpt} and Grounding DINO~\cite{liu2023grounding}. They select pivot frames via GPT, detect objects using Grounding DINO, and then pick specific bounding boxes that best match the given language expression with GPT again.
Consequently, these methods struggle to leverage video-level context and capture the object-level details necessary for understanding complex motion and inter-object distinctions.

%% file: sec/3_method.tex
\section{Method}

\subsection{Overview}
For given $T$ frames of video clip $\mathcal{V}=\{{{I}^t}\}^{T}_{t=1}$, each frame $I^{t}\in\mathbb{R}^{C \times H \times W}$ has height $H$, width $W$, and $C$ channels. In RVOS, a language expression is provided as additional input, and the text encoder tokenizes it into text tokens $\mathcal{E} \in \mathbb{R}^{N_w \times D}$, where $N_w$ denotes the number of tokenized words.
The objective of RVOS is to generate binary mask tracks $\mathcal{B}=\{{B}^{t}\}^{T}_{t=1}$, where each mask ${B}^{t}\in \{0,1\}^{{H}\times{W}}$ corresponds to the referred objects at time $t$.

To address this, we propose a novel framework \ours, which, for the first time, leverages SAM2 object token to efficiently bridge SAM2's knowledge with language features, as illustrated in Figure~\ref{fig:main_figure}.
Specifically, we first generate $N$ candidate mask tracks $\mathcal{M} \in \{0,1\}^{N \times T \times H \times W}$ and their corresponding object tokens $\mathcal{O} \in \mathbb{R}^{N \times T \times D}$ using SAM2, where $D$ denotes the feature dimension. Next, we select the $N_v$ valid mask tracks $\mathcal{M}_{v} \in \mathbb{R}^{N_v \times T \times H \times W}$ that align with the given expression from the candidates $\mathcal{M}$ through our lightweight \textit{language-aligned track selection module}. This module efficiently bridges the modality gap between the object token representations of SAM2 and language features. Through this process, our framework obtain high-quality object mask tracks that are precisely aligned with complex natural language expressions.

\subsection{Preliminary - SAM2}
In this section, we provide an overview of the Segment anything model 2 (SAM2)~\cite{ravi2024sam2}, which is a promptable video segmentation model that consists of an image encoder, a prompt encoder, a mask decoder, and a memory encoder.

\paragrapht{Image encoder.}
The image encoder extracts high-resolution image embeddings from individual frames. These spatial features retains detailed object and scene information. The embeddings are later conditioned on user prompts and past memory for mask generation.

\paragrapht{Prompt encoder.}
The prompt encoder supports three types of user inputs: points $\mathcal{P}_g$, bounding boxes $\mathcal{P}_b$, and masks $\mathcal{P}_m$. It generates prompt tokens representing user inputs that specify the target object for segmentation.

\paragrapht{Mask decoder.}
The mask decoder takes memory-conditioned image embeddings from the memory attention layer and prompt tokens from the prompt encoder as inputs. It generates three mask predictions, each paired with a predicted Intersection over Union (IoU) score and an output mask token. These mask tokens serve as memory values. The final mask is selected based on the highest IoU score, and its associated token is converted into an object pointer $\mathcal{O}^{i,t} \in \mathbb{R}^{D}$ at time $t$ to update the memory for ${i=1,\dots,N}$ and ${t=1,\dots,T}$.

\paragrapht{Memory module.}
SAM2 incorporates a memory module that conditions the features of the current frame on both previous frames and user-provided prompts. Each memory entry consists of two elements: the spatial embedding fused with the predicted mask and a corresponding mask token. By cross-attending to this memory, the model effectively captures fine-grained correspondences and spatial information, ensuring temporal consistency across frames.

\subsection{Track generation}
\label{track_generation}

As our method select the valid mask tracks among the candidates, we first prompt SAM2 to ensure it generates all the objects exist in a video. Since some objects only appear momentarily, we adopt a strategy of selecting frames at predefined frame intervals as a prompt frame $I_p$ for mask generation.

\paragrapht{Prompt mask generation.}
We use two types of input prompts: grid points $\mathcal{P}_g$, and bounding boxes $\mathcal{P}_b$, along with frame $I_p$. The bounding boxes are obtained from external object detection models, only for inference to efficiently capture potential objects.
These prompts are used to generate $N$ binary masks $M_p \in \{0,1\}^{N \times H \times W}$, as
\begin{equation}
    M_p = \text{SAM2}_{\text{Image}}(I_p ; \{\mathcal{P}_g, \mathcal{P}_b\}),
\end{equation}
where $\text{SAM2}_{\text{Image}}(\cdot)$ denotes the SAM2 image predictor.

\paragrapht{Mask track propagation.}
The generated masks $M_p$ are propagated both forward and backward across the entire video $\mathcal{V}$ by the SAM2 video predictor $\text{SAM2}_{\text{Video}}(\cdot)$, to obtain mask tracks $\mathcal{M}$ and the corresponding object tokens $\mathcal{O}$:
\begin{equation}
    \mathcal{O}, \mathcal{M}= \text{SAM2}_{\text{Video}}(\mathcal{V} ;M_p).
\end{equation}
Notably, grid point prompts $\mathcal{P}_g$ cover both the foreground objects and the surrounding background, as the points are evenly distributed across the frame.

\subsection{Object representation}
Revisiting the architecture of SAM2~\cite{ravi2024sam2}, the object pointer obtained from the spatiotemporal-aware memory bank serves as an auxiliary high-level representation of the objects to be segmented.
Specifically, each object pointer $\mathcal{O}^{i,t}$, corresponding to a mask $\mathcal{M}^{i,t}$ within a frame, is hypothesized to encode certain object-level information at that timestep $t$. Consequently, the sequence of these object pointers accumulated over time can be considered as temporal-aware object information, which inherently captures \textit{object motion}.
Based on this intuition, as we generate $N$ candidate mask tracks $\mathcal{M}$ using SAM2, we simultaneously extract $T$ object pointers $\{\mathcal{O}^{i,t}\}_{t=1}^T$ per track $i$ and concatenate them along the temporal dimension. We define the resulting representation as the \textit{object token}, denoted as $\mathcal{O}^i$, which encapsulates both spatial and motion characteristics of the object over time.
Motivated by these considerations, we utilizes these object tokens to model complex motions of objects.

\subsection{Track selection}
\label{track_selection}

Once we successfully gather $N$ candidate mask tracks and their consistent feature representations, $\mathcal{O}$, we can address RVOS by selecting tracks that semantically match the given expression. To determine this, we introduce a lightweight \textit{language-aligned track selection module}, which aligns SAM2's token representations and language features, thus outputs scores reflecting correspondence between each mask track and given language expression. We define these scores as alignment score $s_a$, representing the probability of selection. Thus, the module takes object tokens $\mathcal{O}$ and text tokens $\mathcal{E}$ as input, and produces $s_a\in\mathbb{R}^{N}$ along with an alignment token $\mathcal{O}_a\in\mathbb{R}^{N\times D}$.
\begin{equation}
\mathcal{O}_a, {s}_a =  \text{TS}(\mathcal{O};\mathcal{E}),
\end{equation}
where $\text{TS}(\cdot)$ denotes the track selection module.
As depicted in Figure~\ref{fig:module}, the track selection module is composed of short-term motion encoder followed by object-language alignment layers and language aligned motion aggregation module.

\paragrapht{Short-term motion encoder.}
Since RVOS deals with video data, target objects are not limited to appearance cues; they are often defined by key motion cues. Thus, vision-language alignment in RVOS requires not only frame-level object features but also temporal encoding.
The initial short term motion encoder is to encode the momentary motions of objects, by implementing 1D convolutional network along temporal dimension of each object token. The output object token is $\mathcal{O}^i \in \mathbb{R}^{T' \times D}$, where $T'$ denotes the reduced temporal dimension.

\paragrapht{Object-language alignment layer.}
The object-language alignment layer, repeats $L$ times, sequentially performs three types of attention layers: inter-object attention, motion attention, and object-to-language attention.

Understanding an object's motion implies both its interactions with the surrounding environment and its internal dynamicity. We address these temporal and spatial contexts using motion attention and inter-object attention, respectively. Both attentions are standard self-attention~\cite{vaswani2017attention}, but each operate along a different dimension for distinct pursuit.

Inter-object attention is applied to all object tokens $\mathcal{O}^t \in \mathbb{R}^{N \times D}$ within the same frame $t$. As we aforementioned in Section \ref{track_generation}, using grid point prompts allows us to obtain mask tracks correspond to both foreground and background regions. Thus inter-object attention between all these tokens captures both object relations and interactions between objects and surroundings, leading to a comprehensive understanding of the global context. On the other hand, motion attention aims to aggregate long-term motion information for each object throughout the video, operating along the temporal dimension of each object token $\mathcal{O}^i \in \mathbb{R}^{T' \times D}$.

Subsequently, we employ object-to-language cross-attention to align visual object tokens $\mathcal{O}$ with language features  $\mathcal{E}$, generating language-aligned object token $\mathcal{O}' \in \mathbb{R}^{N \times T' \times D}$. Finally, these alignment tokens $\mathcal{O}'$ serves a input to the language-aligned object aggregation block for further processing.

\input{fig/module}

\paragrapht{Language-aligned object aggregation.}
The language-aligned object aggregation block takes the language-aligned object token $\mathcal{O}'$ as input and outputs $ {s}_a$ along with $\mathcal{O}_a$ which serves as the object representative.
We define $\mathcal{O}_a\in\mathbb{R}^{N \times D}$ as the weighted sum of object tokens, computed using the frame weighting matrix $w_a \in \mathbb{R}^{N \times T'}$, given by:
\begin{equation}
    \label{eq:score_weight}
    w_a = \text{softmax}(\text{Avg}_{N_w}(\mathcal{O}' \mathcal{E}^\text{T})),
\end{equation}
where $\text{Avg}_{N_w}(\cdot)$ represents the mean along the $N_w$ dimension. Using such operation, we can obtain $\mathcal{O}_a$ and $s_a$ as follows:
\begin{equation}
\begin{split}
    \mathcal{O}_a &= \text{Avg}_{T}(w_a \otimes \mathcal{O}'), \\
     {s}_a &= \text{sigmoid}(\text{Avg}_{T}(\text{Avg}_{N_w}(\mathcal{O}'\mathcal{E}^\text{T}))),
\end{split}
\end{equation}
where $\otimes$ denotes element-wise multiplication operation, and $\text{Avg}_{T}(\cdot)$ represents the mean along the $T$ dimension.

Here, $\mathcal{O}_a$, is not only aligned with the language expression but has also incorporated temporally aggregated motion information, making it a rich video-level object representation.
Then, the alignment score of each object $s^i_a$ is mapped to the $[0,1]$ range, following a sigmoid activation. The $i$-th mask track is then selected or discarded based on whether $s^i_a$ exceeds threshold $\tau$. The selected mask tracks $\mathcal{M}_v$ are merged to form the final output binary mask track $\mathcal{B}$ for the given expression.

\subsection{Training objective}

\paragraph{Pseudo labeling.}
As we utilize SAM2 object token representations in a fully frozen state, our goal is to select the correct object tokens, representing SAM2 generated mask tracks, that matches to the referred object in a given expression. However, since RVOS datasets provide only language expressions paired with ground-truth mask tracks, there is no explicit label for each generated mask tracks and corresponding object token. This means that direct supervision for learning alignment between SAM2 object tokens and language features is unavailable in current setting.
To address this issue, we introduce a novel \textit{IoU-based pseudo-labeling} strategy that enables our model to directly identify which generated mask tracks correspond to a given expression.
Specifically, we compute the mean Intersection over Union (mIoU) between each candidate mask track and the ground-truth track associated with that expression. Candidate object tokens exceeding predefined mIoU threshold $\tau$ are labeled as positive samples, while the rest are labeled as negative samples.
The core motivation of this approach is to create a reliable supervision signal that bridges the gap between the frozen SAM2 object tokens and the language expressions. This mIoU-based pseudo-labeling strategy not only provides a clear supervision but also simplifies training to a straightforward binary classification objective.

\paragrapht{Loss functions.}
The total loss $\mathcal{L}$ is a combination of Binary Cross-Entropy (BCE) loss $\mathcal{L}_\text{BCE}$ and alignment loss $\mathcal{L}_\text{align}$: $\mathcal{L}=\lambda_{1}\mathcal{L}_\text{BCE} + \lambda_{2}\mathcal{L}_\text{align}$.

The BCE loss $\mathcal{L}_\text{BCE}$ is applied to enforce alignment between the object features and the language, as follows: 
\begin{equation}
\mathcal{L}_\text{BCE} = -\frac{1}{N} \sum_{i=1}^{N} \Big( y^{i} \log( {s}_a^{i}) \\
+ (1 - y^{i}) \log(1 - {s}_a^{i}) \Big), 
\end{equation} 
where $y^i$ represents the pseudo binary classification label. Alignment score $s_a$ is defined based on whether the mask track $\mathcal{M}^{i}$ of $\mathcal{O}^{i}$ corresponds to the target object designated by the $\mathcal{E}$.

The alignment loss $\mathcal{L}_\text{align}$ is a modified form of contrastive loss, designed to encourage each alignment token $\mathcal{O}^i_a$ to push mismatched sentences away in semantic space, and vice versa. We define the positive anchor $\mathcal{A}_{p}\in\mathbb{R}^D$ as the mean vector of the text tokens $\mathcal{E}$, ensuring semantic closeness between corresponding tokens.
In contrast, the negative anchors $\mathcal{A}_{n}\in\mathbb{R}^{N_{\text{neg}} \times D}$ consists of $N_\text{neg}$ learnable embeddings, which are trained to represent a distinct negative latent space, forcing the tokens to be pushed farther apart.
$\mathcal{L}_\text{align}$ is defined as follows: 
\begin{equation}
    \mathcal{L}_\text{align}=-\frac{1}{N}\sum_{i=1}^{N}\left(y^i\mathcal{L}_\text{pos}(\mathcal{O}_a^i)+(1-y^i)\mathcal{L}_\text{neg}(\mathcal{O}_a^i)\right),
\end{equation}
where 
\begin{equation}
\begin{split}
\mathcal{L}_\text{pos}&=d(\mathcal{O}_a^i,\mathcal{A}_{p}
)-\sum_{j=1}^{N_{\text{neg}}}d(\mathcal{O}_a^i,\mathcal{A}_{{n}}^j),\\
\mathcal{L}_\text{neg}&=d(\mathcal{O}_a^i,\mathcal{A}_{{n}}^{k^*})-d(\mathcal{O}_a^i,\mathcal{A}_{{p}})-\sum_{\substack{j=1, j \neq k^*}}^{N_{\text{neg}}}d(\mathcal{O}_a^i,\mathcal{A}_{{n}}^j).
\end{split}
\end{equation}
Here, the distance function is computed as \( d(\mathbf{x}, \mathbf{y}) = 1 - \cos(\mathbf{x}, \mathbf{y}) \), where \( \cos(\mathbf{x}, \mathbf{y}) \) is the cosine similarity between vectors \( \mathbf{x} \) and \( \mathbf{y} \). The index \( k^* \) represents the closest negative anchor to the alignment token. 

%% file: fig/module.tex
\begin{figure}[t]
    \centering
    \includegraphics[width=\linewidth]{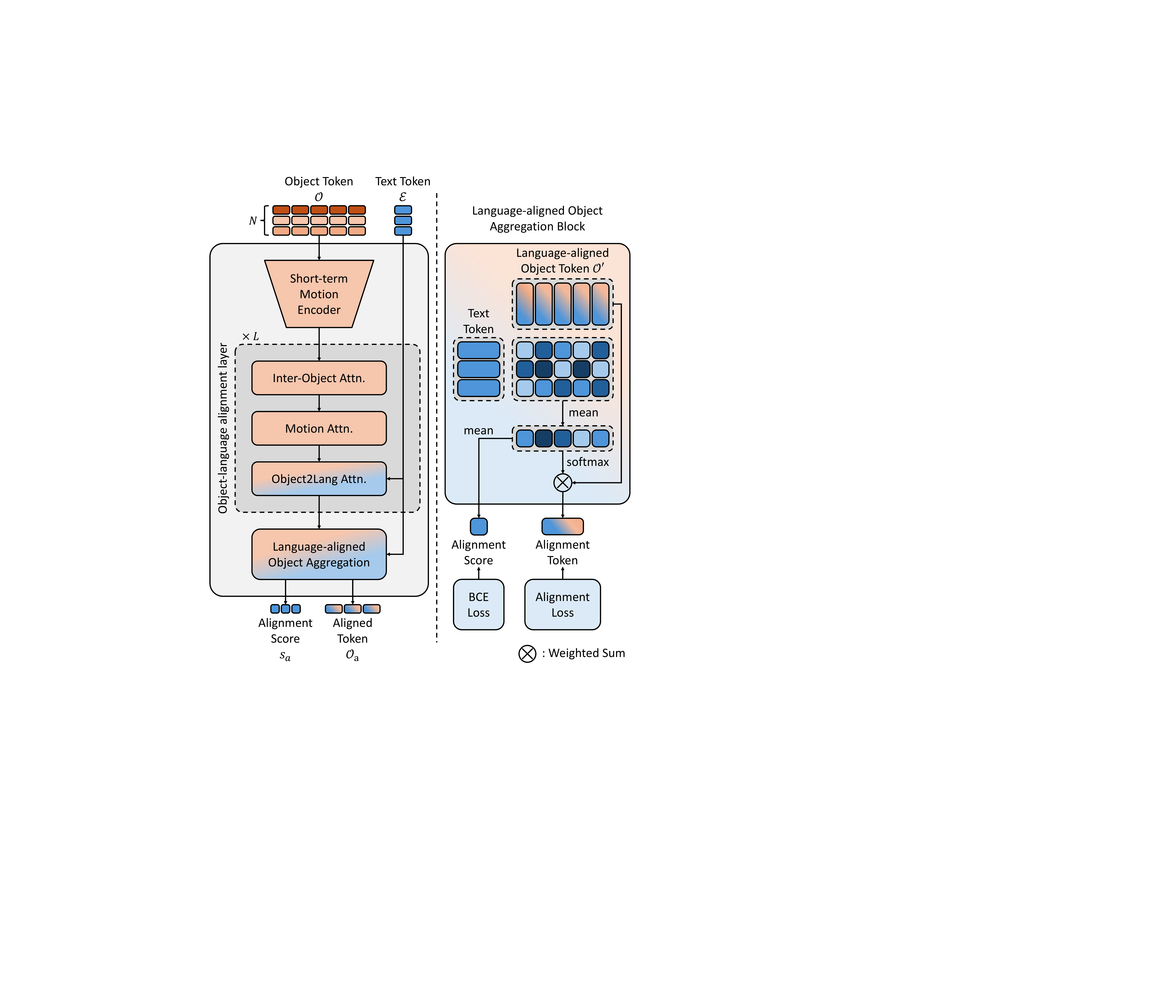}
    \caption{\textbf{Architecture of the language-aligned track selection module}, which takes object tokens and text tokens as inputs, aligning these representations to effectively capture object dynamics.}
    \label{fig:module}
    \vspace{-10pt}
\end{figure}

%% file: sec/4_experiments.tex
\section{Experiments}

\subsection{Datasets and evaluation metrics}

\paragraph{Dataset.}
We evaluate our method on three video datasets: MeViS~\cite{ding2023mevis}, Ref-YouTube-VOS~\cite{seo2020urvos}, and Ref-DAVIS~\cite{KhoRohrSch_ACCV2018}. MeViS, a newly established dataset focused on motion information analysis, comprises 2,006 videos and 28,570 sentences, which are divided into three subsets: the training set with 1,712 videos, the validation set with 140 videos, and the testing set with 154 videos. Ref-YouTube-VOS is the largest RVOS dataset, containing 3,978 videos with approximately 13,000 annotations. Ref-DAVIS builds upon DAVIS17~\cite{perazzi2016benchmark} by incorporating linguistic annotations for a variety of objects, featuring a total of 90 videos.

\paragrapht{Evaluation metrics.}
Following~\cite{ding2023mevis, miao2024htr, he2024decoupling}, we evaluate our method on the MeViS dataset using the commonly used \( \mathcal{J} \)\&\( \mathcal{F} \) metrics. The \( \mathcal{J} \) metric, or region similarity, calculates the Intersection over Union (IoU) between predicted and ground-truth masks to assess segmentation quality, while the \( \mathcal{F} \)-measure evaluates contour accuracy. To provide an overall effectiveness score for our method, we report the average of these two metrics, referred to as \( \mathcal{J} \)\&\( \mathcal{F} \).

\subsection{Implementation details}

\paragrapht{Precomputing SAM2 object tokens.}
Since we utilize SAM2 in a fully frozen state, training focuses exclusively on the language-aligned selection module. Inspired by FuseMix~\cite{vouitsis2024data}, we precompute SAM2 mask tracks on the RVOS dataset, eliminating the need for on-the-fly inference during training. This approach enables efficient training, taking approximately 7 hours on a single RTX 3090 GPU using the MeViS~\cite{ding2023mevis} dataset.

\paragrapht{Track generation.}
We generate mask tracks using SAM2-L~\cite{ravi2024sam2}, prompted by grid points and bounding boxes obtained from Grounding DINO-T~\cite{liu2023grounding} every fourth frame. To avoid generating redundant tracks, we apply IoU-based filtering, similar to Non-Maximum Suppression (NMS)~\cite{neubeck2006efficient}, propagating only distinct prompt masks.

\paragrapht{Language-aligned track selection module.}
We employ pre-trained RoBERTa~\cite{DBLP:journals/corr/abs-1907-11692} as the text encoder. Hyperparameters are set as follows: $N_{\text{neg}}=32$ for number of negative anchors, and loss weights of $\lambda_1=1.0, \lambda_2=0.3$ and $\tau=0.5$ for selection thresholding.

\input{tables/main_quan}
\input{tables/zeroshot_quan}
\input{tables/mevis_youtube_quan}

\subsection{Quantitative results}
\label{sec:quan}

\paragraph{Main results.} Table~\ref{tab:main_quan} presents the quantitative results of our method on the MeViS~\cite{ding2023mevis} dataset, which is widely regarded as the most challenging benchmark in the RVOS field. Our method achieves state-of-the-art performance, underscoring its effectiveness. Additionally, compared to previous methods, \ours significantly reduces the number of trainable parameters to 32.9M while achieving the highest \( \mathcal{J} \)\&\( \mathcal{F} \) score of 48.6. This low number of trainable parameters is achieved through our design, which relies solely exclusively on object tokens.

\paragrapht{Zero-shot evaluation.} Since our method utilizes object tokens obtained from SAM2 in a fully frozen state, we conducted a zero-shot experiment to evaluate its generalization capability. We trained our model on the MeViS~\cite{ding2023mevis} dataset and evaluated it on the Ref-YouTube-VOS~\cite{seo2020urvos} and Ref-DAVIS~\cite{KhoRohrSch_ACCV2018} datasets. As shown in Table~\ref{tab:zeroshot_quan}, \ours achieved superior performance, surpassing the previous state-of-the-art methods. This demonstrates that our approach not only effectively bridges the modality gap between SAM2 token features and language features but also inherits the intrinsic robustness of SAM2 representations.

\paragrapht{Combined dataset evaluation.}
Table~\ref{tab:mevis_youtube_quan} presents the quantitative results obtained by training on a naively combined dataset of MeViS~\cite{ding2023mevis} and Ref-YouTube-VOS~\cite{seo2020urvos}, followed by individual evaluations on each dataset. The results highlight the robustness of our method, as it maintains strong performance across different datasets without requiring dataset-specific tuning. Furthermore, the scalability of our approach is evident, as it effectively leverages multiple datasets without performance degradation, suggesting its potential for broader generalization in RVOS.

\input{tables/ablation}

\subsection{Ablation studies}
\label{ablations}
We conduct our ablation studies on the MeViS~\cite{ding2023mevis} dataset to examine the effectiveness of our proposed language-aligned selection module and its components.

\paragrapht{Effect of the proposed selection method.}
The quantitative results in Table~\ref{tab:abl_selection} demonstrate that our language-aligned track selection module effectively interprets complex language expressions. \textit{w/o selection module} refers to an baseline approach that relies solely on Grounding DINO~\cite{liu2023grounding}, which detects objects at the frame level by understanding the correspondence between text and objects in an image. However, this approach does not incorporate temporal information, limiting its ability to associate objects with motion patterns or temporal events described in the text. Consequently, it struggles with understanding complex expressions that require video-level reasoning. In contrast, \textit{w/ selection module} represents our framework, \ours, which selects the referred object tracks from the candidates by leveraging language-aligned object tokens. By considering both spatial and temporal information, our selection module enables a more comprehensive understanding of complex expressions, leading to improved RVOS.

\paragrapht{Ablation on losses.}
In Table~\ref{tab:abl_loss}, we evaluate the model’s performance under different loss configurations. When using only BCE loss (w/o $\mathcal{L}_\text{align}$), we observe a performance reduction of 4.1  \( \mathcal{J} \)\&\( \mathcal{F} \) compared to the combined setting of BCE and alignment loss (w/ $\mathcal{L}_\text{align}$). This result indicates that alignment loss enhances the model’s discriminative ability, improving its understanding complex motions and enabling more precise alignment with given expression.

\paragrapht{Ablation on different types of attention.}
Table~\ref{tab:abl_attn} shows the model’s performance with different attention configurations. Using only motion attention allows the model to aggregate long-term temporal information across frames, improving motion modeling but neglecting object relationships and scene-level context within each frame. Conversely, using only inter-object attention encodes spatial relationships among objects including surrounding backgrounds, but lacks temporal awareness. Combining both attention types, our method effectively captures temporal object dynamics as well as spatial interactions, resulting in comprehensive global context understanding.

\input{fig/toy_iou_sim}
\input{fig/main_qual}

\subsection{Analysis on object token of SAM2}

As our method solely relies on the object tokens of SAM2, it is important to investigate whether they contain sufficient information to model diverse aspects of corresponding objects. To this end, we conducted two experiments to explore whether these tokens capture motion information from their corresponding masks and whether they possess a minimally sufficient level of semantic knowledge to align with language expressions.

First, we analyzed the relationship between object tokens and their corresponding masks by comparing the cosine similarity of object tokens based on their mIoU (mean Intersection over Union) of their masks. As illustrated in Figure~\ref{fig:toy_iou_sim}, the similarity between tokens tends to increase nearly proportionally to the mIoU. This suggests a correlation between the spatial proximity of masks and the similarity of their object tokens. It provides a reasonable indication that object tokens implicitly encode spatial information, which can be extended to spatial trajectories when temporally connected across frames.

Second, to assess the degree of semantic content encoded in the object tokens, we conducted a simple classification task using the PASCAL-VOC dataset~\cite{everingham2010pascal}, an image dataset containing 20 object categories with pixel-level segmentation and class annotations per mask. Specifically, we added a linear classification head on top of the object tokens and trained it to predict object categories. The classifier achieved a maximum accuracy of $85.3\%$, indicating that the tokens may possess at least a basic ability to differentiate between object classes, and thus contain meaningful semantic information.

These findings suggest that SAM2 object tokens obtain intrinsic properties such as object motion and semantic information. Their potential to encode such \textit{objectness} makes them a promising candidate for aligning with language expressions, offering a lightweight module design.

\subsection{Qualitative results}
In Figure~\ref{fig:main_qual}, our proposed method demonstrates its ability to understand complex language expressions. The model captures both appearance cues—such as \textit{“The cat"} and \textit{“The cow"} attributes—and complex motion cues, including \textit{“moving to right"}.  \ours can select the referred object even when the expression relies solely on motion (e.g., \textit{“chases"}, \textit{“pounces"}).

%% file: tables/main_quan.tex
\renewcommand{\arraystretch}{1.1} 
\setlength{\extrarowheight}{1.3pt} 
\setlength{\aboverulesep}{1.3pt} 
\setlength{\belowrulesep}{1.3pt} 
\setlength{\belowbottomsep}{1.3pt} 
\setlength{\abovetopsep}{1pt} 
\setlength{\lightrulewidth}{0.8pt} 
\begin{table}[t]
    \centering
    \resizebox{0.9\linewidth}{!}{%
    \begin{tabular}{l|c|ccc}
    \toprule
    \multirow{2}{*}{Methods} & \multicolumn{1}{c|}{\# of trainable} & \multicolumn{3}{c}{Metrics} \\
    & \multicolumn{1}{c|}{parameters} & \multirow{1}{*}{\( \mathcal{J} \)\&\( \mathcal{F} \)} & \multirow{1}{*}{\( \mathcal{J} \)} & \multirow{1}{*}{\( \mathcal{F} \)} \\
    \midrule
    \addlinespace[1pt]
    \midrule
    URVOS~\cite{seo2020urvos} & - & 27.8 & 25.7 & 29.9  \\
    LBDT~\cite{ding2022language} & 95.6 M & 29.3 & 27.8 & 30.8  \\
    MTTR~\cite{botach2022end} & - & 30.0 & 28.8 & 31.2  \\
    ReferFormer~\cite{wu2022language} & 112.9 M & 31.0 & 29.8 & 32.2  \\
    VLT+TC~\cite{vision-language-transformer} & \underline{38.3 M} & 35.5 & 33.6 & 37.3  \\
    LMPM~\cite{ding2023mevis} & 66.4 M & 37.2 & 34.2 & 40.2  \\
    HTR~\cite{miao2024htr} & - & 42.7 & 39.9 & 45.5  \\
    DsHmp~\cite{he2024decoupling} & 92.4 M & \underline{46.4} & \underline{43.0} & \underline{49.8}  \\
    \midrule
    \rowcolor{lightgray}
    \textbf{\ours} & \textbf{32.9 M} & \textbf{48.6} & \textbf{45.2} & \textbf{52.1}   \\
    \bottomrule
    \end{tabular}
    }
    \vspace{-5pt}
    \caption{\textbf{Quantitative comparison on MeViS.} The best results are highlighted in \textbf{bold}, and the second-best results are \underline{underlined}.}
    \label{tab:main_quan}
    \vspace{-10pt}
\end{table}

%% file: tables/zeroshot_quan.tex
\setlength{\extrarowheight}{1pt} 
\setlength{\aboverulesep}{1pt} 
\setlength{\belowrulesep}{1pt} 
\setlength{\belowbottomsep}{1pt} 
\setlength{\abovetopsep}{1pt} 
\setlength{\lightrulewidth}{0.8pt} 
\begin{table}[t]
    \centering
    \resizebox{\linewidth}{!}{ 
    \begin{tabular}{l|ccc|ccc}
    \toprule
    & \multicolumn{3}{c|}{Ref-YouTube-VOS} & \multicolumn{3}{c}{Ref-DAVIS} \\
    \multirow{-2}{*}{Methods} & \( \mathcal{J} \)\&\( \mathcal{F} \) & \( \mathcal{J} \) & \( \mathcal{F} \) & \( \mathcal{J} \)\&\( \mathcal{F} \) & \( \mathcal{J} \) & \( \mathcal{F} \) \\
    \midrule
    \addlinespace[1pt]
    \midrule
    ReferFormer~\cite{jia2021scaling} & 35.0 & 34.2 & 35.8 & 40.5 & 36.8 & 44.2 \\
    LMPM~\cite{ding2023mevis} & 31.5 & 30.0 & 32.9 & 39.9 & 36.7 & 43.2 \\
    DsHmp~\cite{he2024decoupling} & 45.8 & 43.7 & 47.9 & 42.6 & 37.8 & 47.3 \\
    \midrule
    \rowcolor{lightgray}
    \textbf{\ours} & \textbf{47.9} & \textbf{44.3} & \textbf{51.5} & \textbf{45.4} & \textbf{43.0} & \textbf{47.7} \\
    \bottomrule
    \end{tabular}
    }
    \vspace{-5pt}
    \caption{\textbf{Zero-shot quantitative comparison on Ref-YouTube-VOS and Ref-DAVIS.} The best results are in \textbf{bold}. The models are trained on the training set of MeViS and evaluated on Ref-YouTube-VOS and Ref-DAVIS.}
    \label{tab:zeroshot_quan}
    \vspace{-10pt}
\end{table}

%% file: tables/mevis_youtube_quan.tex
\setlength{\extrarowheight}{1pt} 
\setlength{\aboverulesep}{1pt} 
\setlength{\belowrulesep}{1pt} 
\setlength{\belowbottomsep}{1pt} 
\setlength{\abovetopsep}{1pt} 
\setlength{\lightrulewidth}{0.8pt} 
\begin{table}[t]
    \centering
    \resizebox{\linewidth}{!}{ 
    \begin{tabular}{l|ccc|ccc}
    \toprule
    & \multicolumn{3}{c|}{MeViS} & \multicolumn{3}{c}{Ref-YouTube-VOS} \\
    \multirow{-2}{*}{Methods} & \( \mathcal{J} \)\&\( \mathcal{F} \) & \( \mathcal{J} \) & \( \mathcal{F} \) & \( \mathcal{J} \)\&\( \mathcal{F} \) & \( \mathcal{J} \) & \( \mathcal{F} \) \\
    \midrule
    \addlinespace[1pt]
    \midrule
    ReferFormer~\cite{jia2021scaling} & 36.6 & 34.1 & 39.1 & 46.8 & 46.2 & 47.5 \\
    LMPM~\cite{ding2023mevis} & 40.5 & 37.8 & 43.2 & 37.6 & 36.0 & 39.2 \\
    DsHmp~\cite{he2024decoupling} & 42.5 & 37.5 & 47.4 & 51.4 & 48.5 & 54.3 \\
    \midrule
    \rowcolor{lightgray}
    \textbf{\ours} & \textbf{48.9} & \textbf{45.2} & \textbf{52.6} & \textbf{55.4} & \textbf{52.0} & \textbf{58.8} \\
    \bottomrule
    \end{tabular}
    }
    \vspace{-5pt}
    \caption{\textbf{Quantitative comparison on combined dataset.} The best results are in \textbf{bold}. The models are jointly trained on the training sets of MeViS and Ref-YouTube-VOS and evaluated separately on their respective evaluation datasets.}
    \label{tab:mevis_youtube_quan}
    \vspace{-10pt}
\end{table}

%% file: tables/ablation.tex
\renewcommand{\arraystretch}{1.1} 
\setlength{\heavyrulewidth}{0.8pt} 
\setlength{\lightrulewidth}{0.8pt} 
\setlength{\cmidrulewidth}{0.5pt}  
\setlength{\extrarowheight}{1pt} 
\setlength{\aboverulesep}{1pt} 
\setlength{\belowrulesep}{1pt} 
\setlength{\belowbottomsep}{1pt} 
\setlength{\abovetopsep}{1pt} 

\begin{table}[t]
    \centering
    \resizebox{0.35\textwidth}{!}{%
    \smaller[0.8]{
        \begin{tabular}{c|ccc}
            \toprule
            Methods & \( \mathcal{J}\&\mathcal{F} \) & \( \mathcal{J} \) & \( \mathcal{F} \) \\
            \midrule
            \addlinespace[1pt]
            \midrule
            w/o selection module & 36.9 & 30.0 & 43.8 \\
            \rowcolor{lightgray}
            w/ selection module & \textbf{48.6} & \textbf{45.2} & \textbf{52.1} \\
            \bottomrule
        \end{tabular}%
    }}
    \vspace{-5pt}
    \caption{\textbf{Ablation study on our proposed selection method.}}
    \label{tab:abl_selection}
    \vspace{-10pt}
\end{table}

\begin{table}[t]
    \centering
    \small 
    
    \begin{minipage}{\columnwidth}
        \centering
        \begin{tabular}{c|ccc}
            \toprule
            Methods & \( \mathcal{J}\&\mathcal{F} \) & \( \mathcal{J} \) & \( \mathcal{F} \) \\
            \midrule
            \addlinespace[1pt]
            \midrule
            w/o $\mathcal{L}_\text{align}$ & 44.5 & 41.4 & 47.6 \\
            \rowcolor{lightgray}
            w/ $\mathcal{L}_\text{align}$ & \textbf{48.6} & \textbf{45.2} & \textbf{52.1} \\
            \bottomrule
        \end{tabular}
        \subcaption{Different loss functions.}
        \label{tab:abl_loss}
    \end{minipage}
    
    \begin{minipage}{\columnwidth}
        \centering
        \begin{tabular}{c|c|ccc}
            \toprule
            Inter-object attn. & Motion attn. & \( \mathcal{J}\&\mathcal{F} \) & \( \mathcal{J} \) & \( \mathcal{F} \) \\
            \midrule
            \addlinespace[1pt]
            \midrule
            \xmark & \cmark & 44.3 & 41.6 & 47.0 \\
            \cmark & \xmark & 44.9 & 42.2 & 47.0 \\
            \rowcolor{lightgray}
            \cmark & \cmark & \textbf{48.6} & \textbf{45.2} & \textbf{52.1} \\
            \bottomrule
        \end{tabular}
        \subcaption{Effects of employing different types of attention layers.}
        \label{tab:abl_attn}
    \end{minipage}
    \vspace{-5pt}
    \caption{\textbf{Ablation studies on various settings of our method.}}
    \vspace{-10pt}
\end{table}

%% file: fig/toy_iou_sim.tex
\begin{figure}[t]
    \centering
    \includegraphics[width=0.95\linewidth]{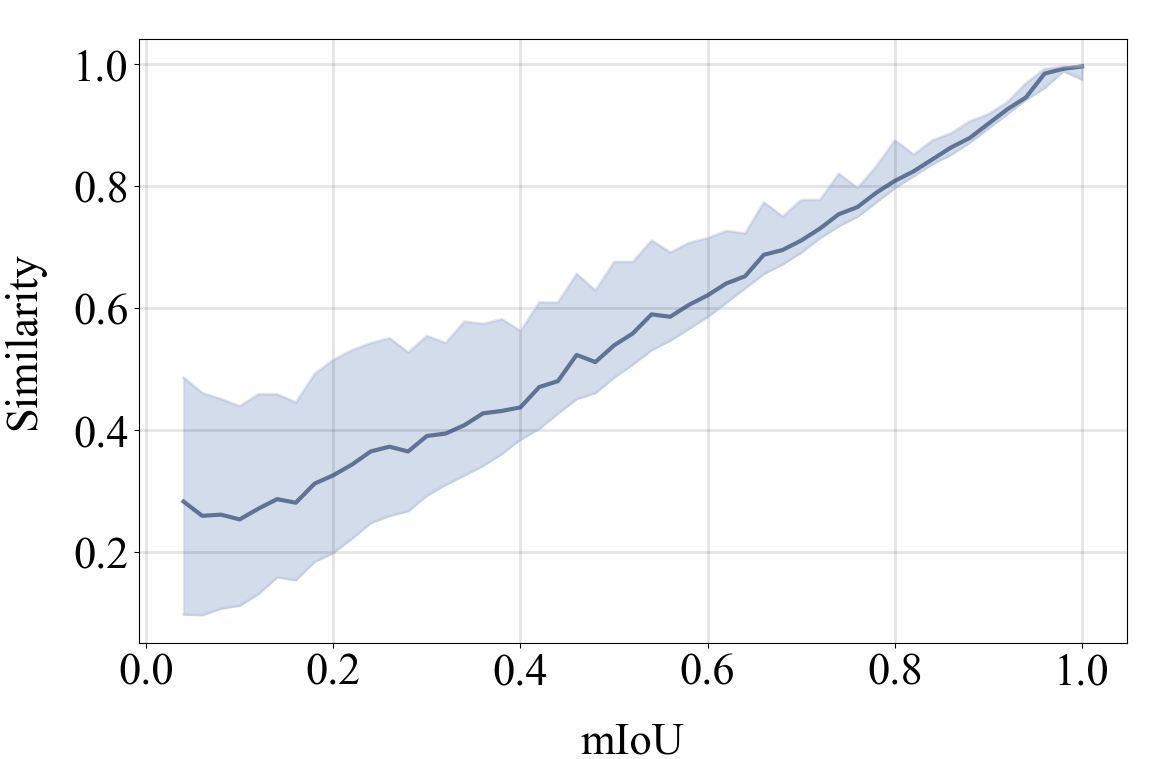}
    \vspace{-5pt}
    \caption{\textbf{Spatial and motion information in object tokens.} The bold line represents the mean similarity, while the shaded region indicates the variance. The results show a certain correlation: as the mIoU between mask tracks increases, the similarity between their associated tokens also rises nearly proportionally. This tendency suggests that object tokens inherently capture spatial information, implicitly encoding object motions over time.}
    \label{fig:toy_iou_sim}
    \vspace{-10pt}
\end{figure}

%% file: fig/main_qual.tex
\begin{figure*}[htbp]
    \centering
    \includegraphics[width=0.8\linewidth]{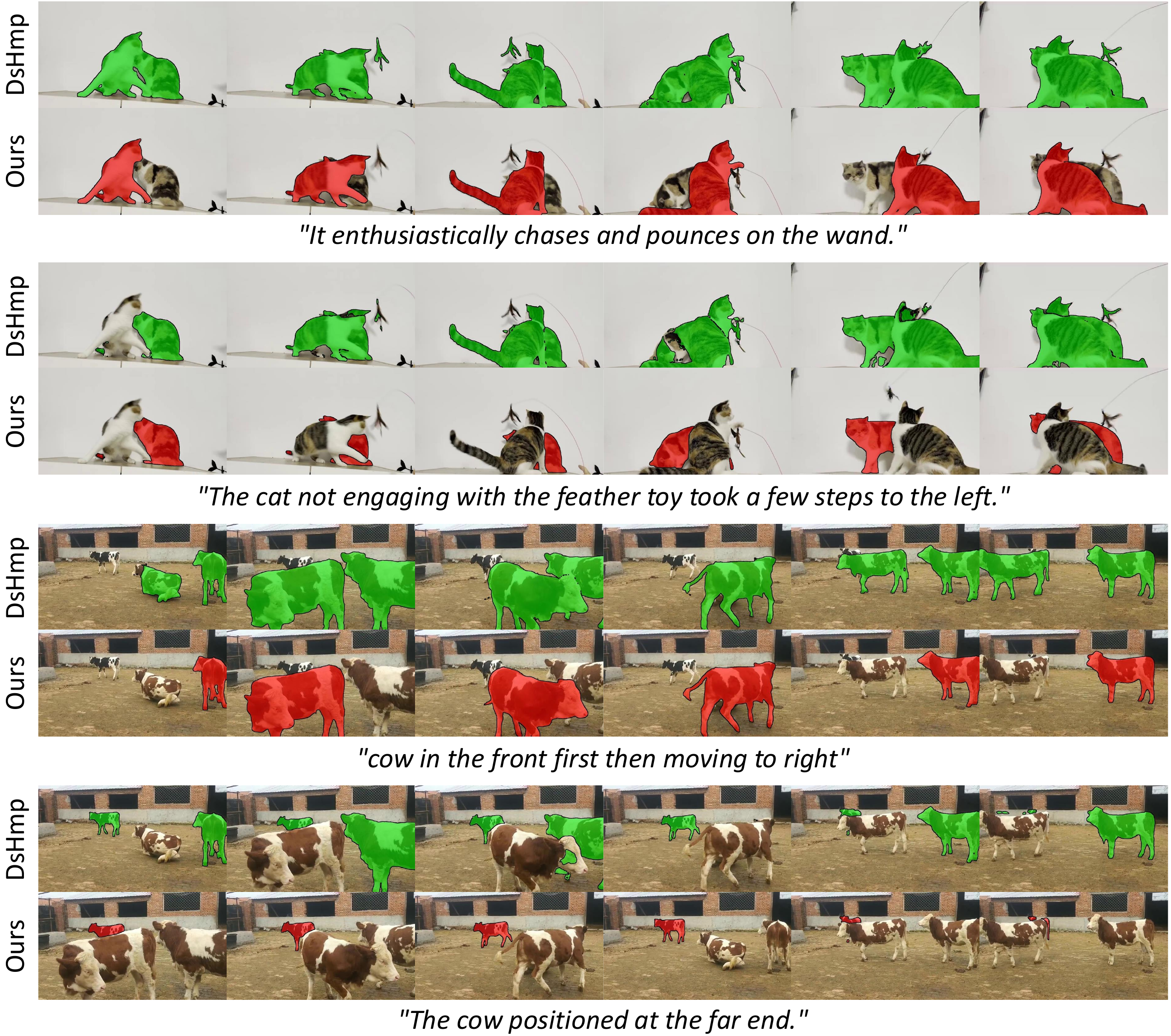}
    \vspace{-5pt}
    \caption{\textbf{Qualitative results of our model on MeViS.} \ours shows its ability to understand complex language expressions.}
    \label{fig:main_qual}
    \vspace{-10pt}
\end{figure*}

%% file: sec/5_conclusion.tex
\section{Conclusion and discussion}

\label{conclusion}
We proposed \ours, a novel framework that leverages SAM2 object tokens as compact video-level object representation.
We align these object tokens with language features using a lightweight track selection module with only 32.9M trainable parameters. Additionally, we employ an IoU-based pseudo-labeling strategy to effectively bridge the modality gap between SAM2 representations and language features.
Our experiments demonstrate that \ours achieves state-of-the-art results on the MeViS dataset.
This validates the effectiveness of \ours in addressing the challenges of complex motion understanding and multi-modal alignment in RVOS.

%% file: supplementary.tex
\renewcommand{\thesection}{\Alph{section}}
\renewcommand{\thefigure}{A.\arabic{figure}}
\renewcommand{\thetable}{A.\arabic{table}}
\setcounter{section}{0}
\setcounter{figure}{0}
\setcounter{table}{0}

\maketitlesupplementary

\setcounter{page}{1}

\section{Additional qualitative results}
\label{suppl:qual}

\paragraph{Qualitative results on MeViS.}
Figure~\ref{fig:suppl_mevis_qual} presents the qualitative results on MeViS~\cite{ding2023mevis}, comparing the performance of DsHmp~\cite{he2024decoupling} with \ours. Our approach consistently demonstrates superior capability in accurately selecting the target object as specified by the referring expression.
Specifically, Figure~\ref{fig:suppl_mevis_two_exp_qual} illustrates scenarios involving a single video with two distinct expressions. \ours accurately identifies the precise object corresponding to each expression, whereas DsHmp demonstrates limitations in distinguishing between objects described by different expressions. Figure~\ref{fig:suppl_mevis_only_motion} illustrates a scenario where the given expression exclusively describes motion-related information (e.g., \textit{``Going right."}). Our language-aligned track selection module can establish correspondence with the expression using motion cues from the language alone, independently of appearance-based features.
 
\paragrapht{Qualitative results on Ref-YouTube-VOS.}
Figure~\ref{fig:suppl_ytbvos_qual} presents the qualitative results on the Ref-YouTube-VOS~\cite{seo2020urvos} dataset in a zero-shot setting, where the model has been trained on MeViS dataset. The results highlights our model's remarkable capability to generalize across diverse videos and expressions, despite not having seen the dataset during training.
This generalization underscores the strength of our approach in leveraging the intrinsic robustness of SAM2 representations.

\section{Results on corrupted setting}
To demonstrate the robustness of our method, we evaluated it on a perturbed dataset with ImageNet-C~\cite{hendrycks2019benchmarking} derived corruption. we intentionally corrupted all video frames with gaussian noise or motion blur, simulating common distortions in real-world scenarios such as low-light environments or rapid camera movements. Since these perturbations represent data types not originally present in the dataset, our method’s ability to effectively handle them shows its robustness inherited from SAM2 and highlights its suitability for practical applications. Table~\ref{tab:noise_quan} presents the quantitative results, showing that our proposed method outperforms previous approaches~\cite{ding2023mevis, he2024decoupling, jia2021scaling}  even under corruption scenarios.

\input{suppl/tables/noise_quan}

\paragrapht{Qualitative results on MeViS with image corruption.}
Figures~\ref{fig:suppl_gaussian_qual} and~\ref{fig:suppl_motion_qual} visualize the results presented in Table~\ref{tab:noise_quan}. These results demonstrate that \ours consistently retains its ability to select the correct object even in corrupted environments.

\section{Additional ablation studies on MeViS}

\paragrapht{Existence of background object tokens.}
The quantitative results presented in Table~\ref{tab:abl_background} underscore the effectiveness of incorporating background object tokens during both training and inference. During training, background object tokens refer to object tokens corresponding to mask tracks that have low IoU with the ground-truth mask track, while during inference, they are derived from mask tracks obtained using grid point prompts. Given that the inter-object attention is designed to capture object relationships and scene-level understanding, the inclusion of background object tokens in both training and inference significantly enhances performance. This comprehensive interactions between foreground and background objects proves its effectiveness, enabling a more enhanced video-level understanding of language.

\paragrapht{Ablation on the number of object-language alignment layers.}
Table~\ref{tab:abl_layer} shows the results of using different numbers of attention block layers.
Our method achieves the highest performance when two layers are adopted, compared to the settings with one or three layers.

\input{suppl/tables/abl}

\section{Detailed implementation details}
\paragrapht{Precomputing SAM2 object tokens.}
Since our method operates with a fully frozen SAM2 and trains only the language-aligned selection module using object tokens, we adopt a highly efficient training strategy similar to FuseMix~\cite{vouitsis2024data}. Specifically, we first perform SAM2 mask propagation on the given RVOS dataset to generate candidate mask tracks and their corresponding object tokens in advance. By precomputing these tokens beforehand, we eliminate the need for SAM2 inference during training phase, allowing us to focus solely on optimizing the language-aligned track selection module. The entire training process, using the MeViS~\cite{ding2023mevis} training dataset, takes approximately 7 hours on a single RTX 3090 GPU.

\paragrapht{Track generation.}
We employ grid points and bounding boxes from the object detection model, Grounding DINO (GDINO)-T~\cite{liu2023grounding} every fourth frame to generate prompt masks, which serve as input for the SAM2-L~\cite{ravi2024sam2} video predictor.
To reduce redundant mask track generation, we filter out similar prompt masks based on their Intersection over Union (IoU) scores. Specifically, we first propagate the mask track sequence starting from the largest prompt mask. Then, for each subsequent prompt mask, we filter it out if its IoU with the previously generated mask tracks at the corresponding frame exceeds 0.7, ensuring that only distinct prompt masks propagate new tracks.

\paragrapht{Language-aligned track selection module.}
We employ pre-trained RoBERTa~\cite{DBLP:journals/corr/abs-1907-11692} as the text encoder. Training is conducted over 13 epochs, with an initial learning rate of 5e-6 that gradually decreases throughout training. We set the hyperparameter values for $\lambda_1$, $\lambda_2$, $N_{\text{neg}}$, $\tau$ to 1.0, 0.3, 32, and 0.5, respectively.

\section{Limitations and future works}
While our approach effectively solve RVOS, certain aspects remain beyond the scope of our work.
The training objectives of the text encoder and the RVOS model differ: the text encoder is trained to identify the best matching words from the vocabulary, while the RVOS model focuses on extracting key cues from sentences essential for locating the corresponding objects. In our future work, we aim to explore tuning the text encoder to capture features that are particularly beneficial for the RVOS task.

\input{suppl/fig/mevis_qual}
\input{suppl/fig/mevis_two_exp_qual}
\input{suppl/fig/mevis_only_motion}

\clearpage
\input{suppl/fig/ytbvos_qual}

\input{suppl/fig/noise_gaussian_qual}
\input{suppl/fig/noise_motion_qual}

\clearpage
\newpage

\bigskip
\clearpage
\newpage

%% file: suppl/tables/noise_quan.tex
\setlength{\extrarowheight}{1pt}
\setlength{\aboverulesep}{1pt}
\setlength{\belowrulesep}{1pt}
\setlength{\belowbottomsep}{1pt}
\setlength{\abovetopsep}{1pt}
\setlength{\lightrulewidth}{0.5pt}
\begin{table}[t]
    \centering
    \resizebox{\columnwidth}{!}{%
    \begin{tabular}{l|c|ccc}
    \toprule
    Methods & Algorithm & \( \mathcal{J} \)\&\( \mathcal{F} \) & \( \mathcal{J} \) & \( \mathcal{F} \) \\
    \midrule
    \addlinespace[1pt]
    \midrule
    ReferFormer~\cite{jia2021scaling} & \multirow{4}{*}{\parbox[c]{2cm}{\centering Motion blur}} & 26.3 & 25.4 & 27.1 \\
    LMPM~\cite{ding2023mevis} &  & 33.3 & 31.2 & 35.4 \\
    DsHmp~\cite{he2024decoupling} &  & 38.0 & 35.0 & 41.1 \\
    \cellcolor{lightgray}\textbf{\ours} &  & \cellcolor{lightgray}\textbf{39.8} & \cellcolor{lightgray}\textbf{36.6} & \cellcolor{lightgray}\textbf{43.0} \\
    \midrule
    ReferFormer~\cite{jia2021scaling} & \multirow{4}{*}{\parbox[c]{2cm}{\centering Gaussian noise}} & 26.9 & 24.0 & 29.9 \\
    LMPM~\cite{ding2023mevis} &  & 36.0 & 33.4 & 38.6 \\
    DsHmp~\cite{he2024decoupling} &  & 43.4 & 39.5 & 47.2 \\
    \cellcolor{lightgray}\textbf{\ours} &  & \cellcolor{lightgray}\textbf{44.4} & \cellcolor{lightgray}\textbf{40.5} & \cellcolor{lightgray}\textbf{48.3} \\
    \bottomrule
    \end{tabular}%
}
    \caption{\textbf{Quantitative result on a corrupted version of MeViS.} The best results are in \textbf{bold}. The models are trained on the original training set and evaluated on the corrupted version of the validation set. The image corruption algorithms are derived from ImageNet-C~\cite{hendrycks2019benchmarking}, with corruption severity 5.}
    \label{tab:noise_quan}
\end{table}

%% file: suppl/tables/abl.tex
\setlength{\extrarowheight}{1pt}
\setlength{\aboverulesep}{1pt}
\setlength{\belowrulesep}{1pt}
\setlength{\belowbottomsep}{1pt}
\setlength{\abovetopsep}{1pt}

\begin{table}[t]
    \centering
    \small
    \begin{minipage}{\columnwidth}
        \centering
        \begin{tabular}{c|c|ccc}
            \toprule
            Train & Inference & \( \mathcal{J}\&\mathcal{F} \) & \( \mathcal{J} \) & \( \mathcal{F} \) \\
            \midrule
            \addlinespace[1pt]
            \midrule
            \xmark & \xmark & 45.7 & 42.4 & 48.9 \\
            \cmark & \xmark & 47.5 & 43.9 & 51.1 \\
            \rowcolor{lightgray}
            \cmark & \cmark & \textbf{48.6} & \textbf{45.2} & \textbf{52.1} \\
            \bottomrule
        \end{tabular}
        \subcaption{Effects of including background object tokens.}
        \label{tab:abl_background}
    \end{minipage}
    
    \begin{minipage}{\columnwidth}
        \centering
        \begin{tabular}{c|ccc}
            \toprule
            \# of Alignment Layers & \( \mathcal{J}\&\mathcal{F} \) & \( \mathcal{J} \) & \( \mathcal{F} \) \\
            \midrule
            \addlinespace[1pt]
            \midrule
            1 & 42.5 & 40.0 & 45.1 \\
            \rowcolor{lightgray}
            2 & \textbf{48.6} & \textbf{45.2} & \textbf{52.1} \\
            3 & 48.2 & 44.8 & 51.5 \\
            \bottomrule
        \end{tabular}
        \subcaption{Effects of the number of object-language alignment layers.}
        \label{tab:abl_layer}
    \end{minipage}

    \caption{\textbf{Additional ablation studies on various settings of our method.}}
\end{table}

%% file: suppl/fig/mevis_qual.tex
\begin{figure*}[ht]
    \centering
    \vspace{-10pt}
    \includegraphics[width=\linewidth]{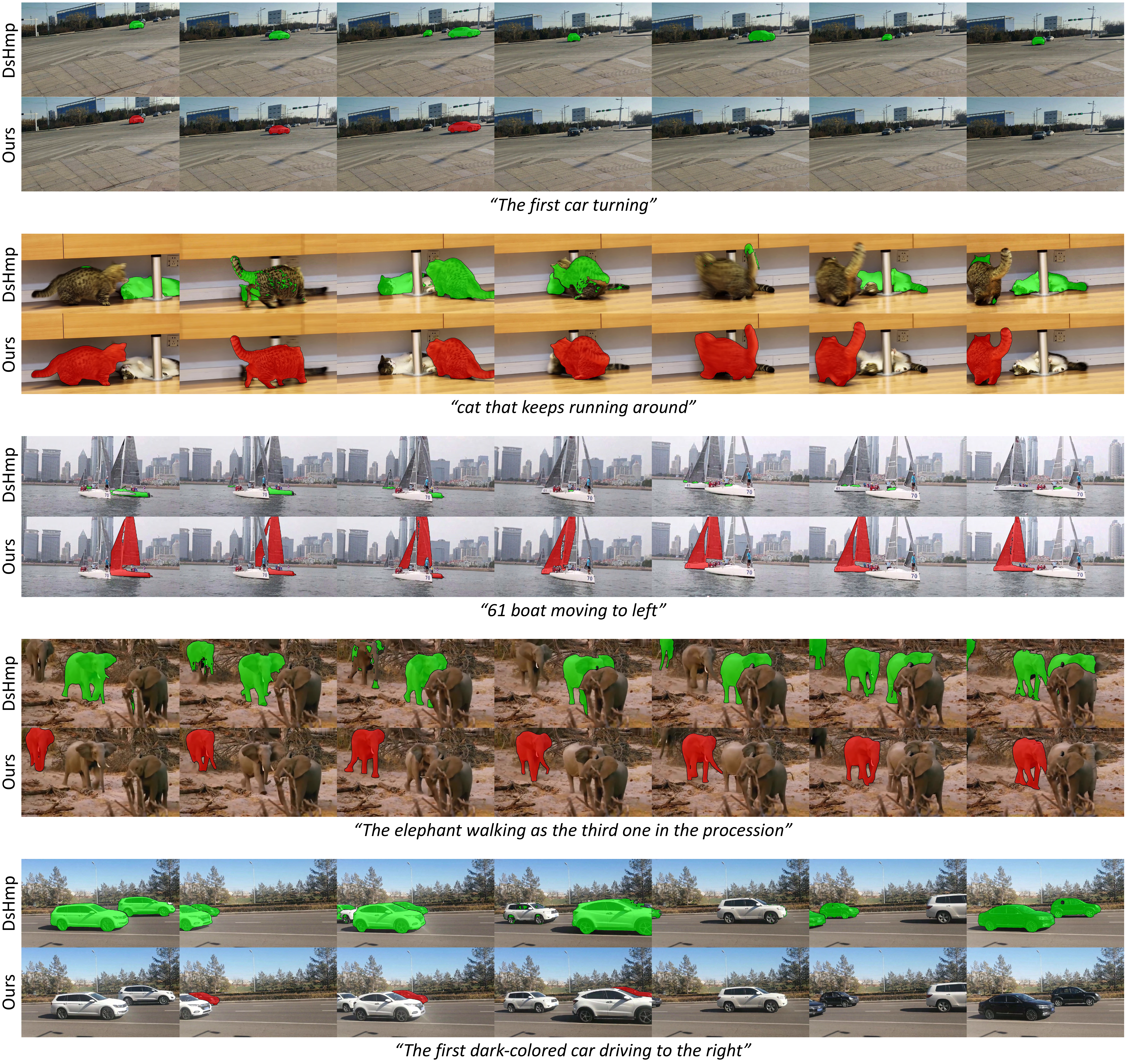}
    \vspace{-15pt}
\caption{\textbf{Qualitative results on MeViS.} Our proposed method outperforms previous state-of-the-art approach~\cite{he2024decoupling} in terms of mask quality and tracking ability, while ensuring accurate segmentation of the corresponding object based on the given expression.}
    \label{fig:suppl_mevis_qual}
\end{figure*}

%% file: suppl/fig/mevis_two_exp_qual.tex
\begin{figure*}[ht]
    \centering
    \vspace{-10pt}
    \includegraphics[width=\linewidth]{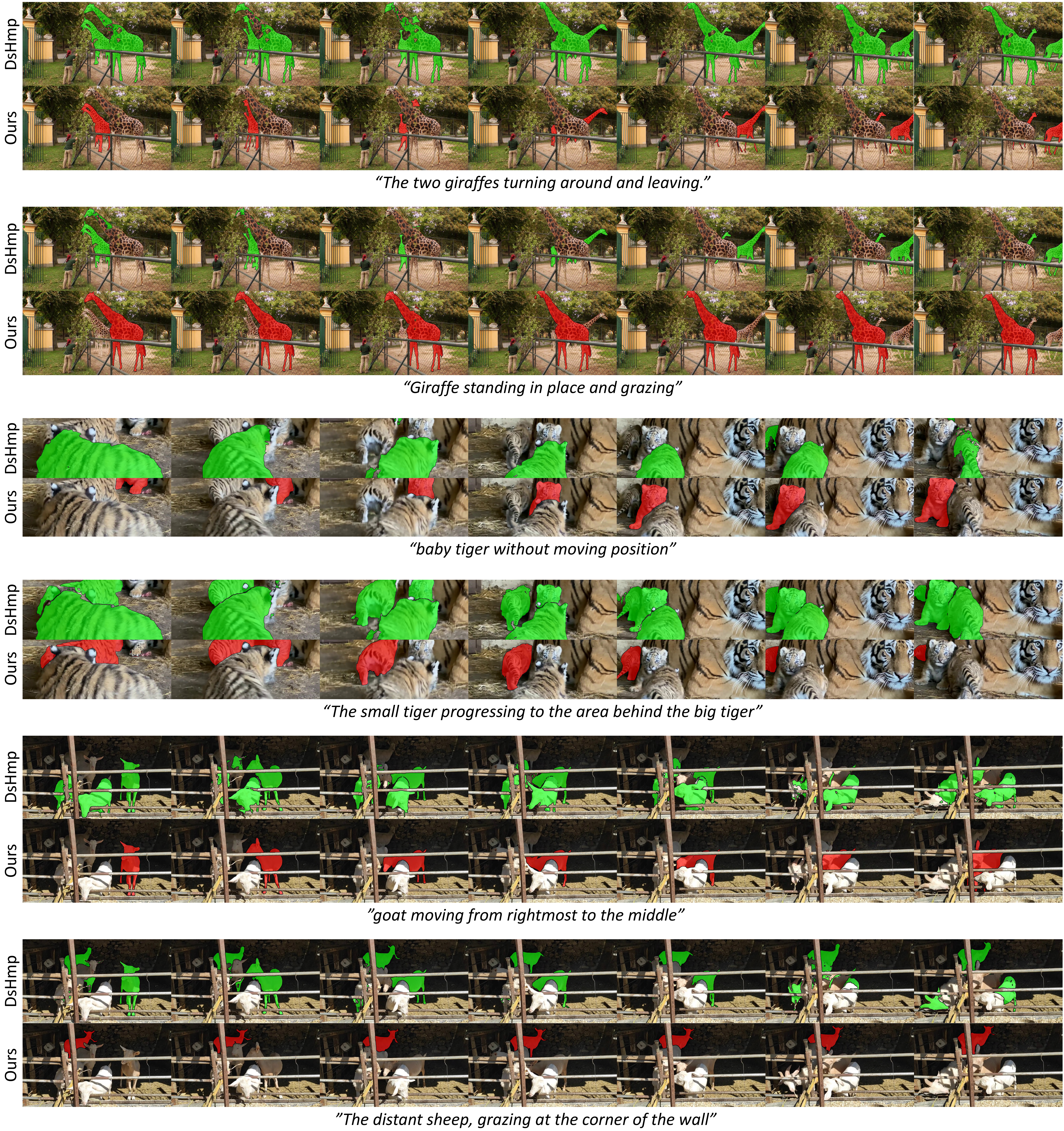}
    \vspace{-15pt}
\caption{\textbf{Qualitative results on MeViS.} Our proposed method outperforms previous state-of-the-art approach~\cite{he2024decoupling} in terms of accurate selection of the corresponding object, while ensuring accurate segmentation of the corresponding object based on the given expression.}
    \label{fig:suppl_mevis_two_exp_qual}
\end{figure*}

%% file: suppl/fig/mevis_only_motion.tex
\begin{figure*}[t]
    \centering
    \vspace{-10pt}
    \includegraphics[width=\linewidth]{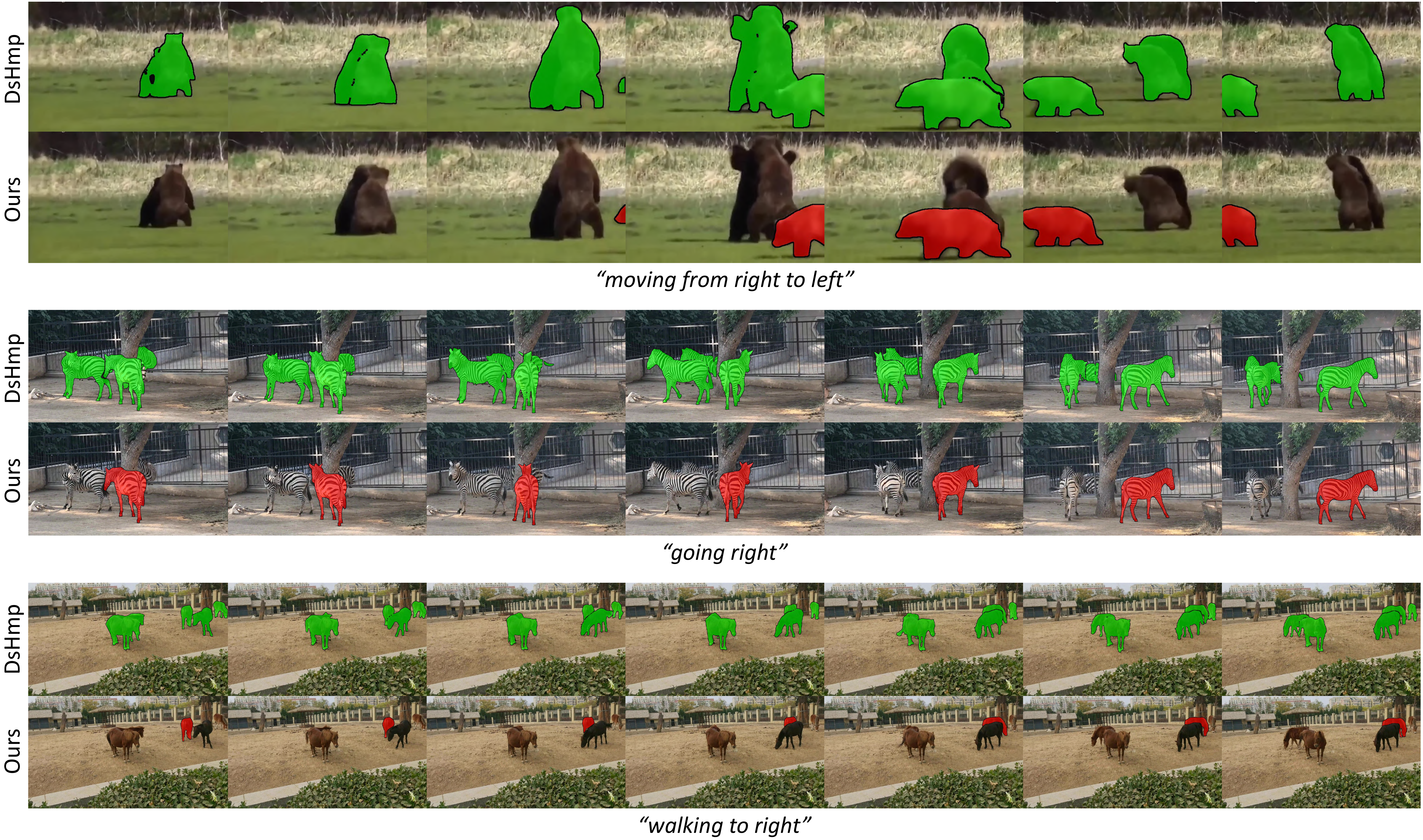}
\caption{\textbf{Qualitative results on MeViS.} Our proposed method outperforms previous state-of-the-art approach~\cite{he2024decoupling} in terms of accurate selection of the corresponding object, while ensuring accurate segmentation of the corresponding object based on the given expression. Notably, despite the given expression focusing solely on motion information, our model effectively handles the task without relying on appearance cues.}
    \label{fig:suppl_mevis_only_motion}
\end{figure*}

%% file: suppl/fig/ytbvos_qual.tex
\begin{figure*}[t]
    \centering
    \vspace{-10pt}
    \includegraphics[width=\linewidth]{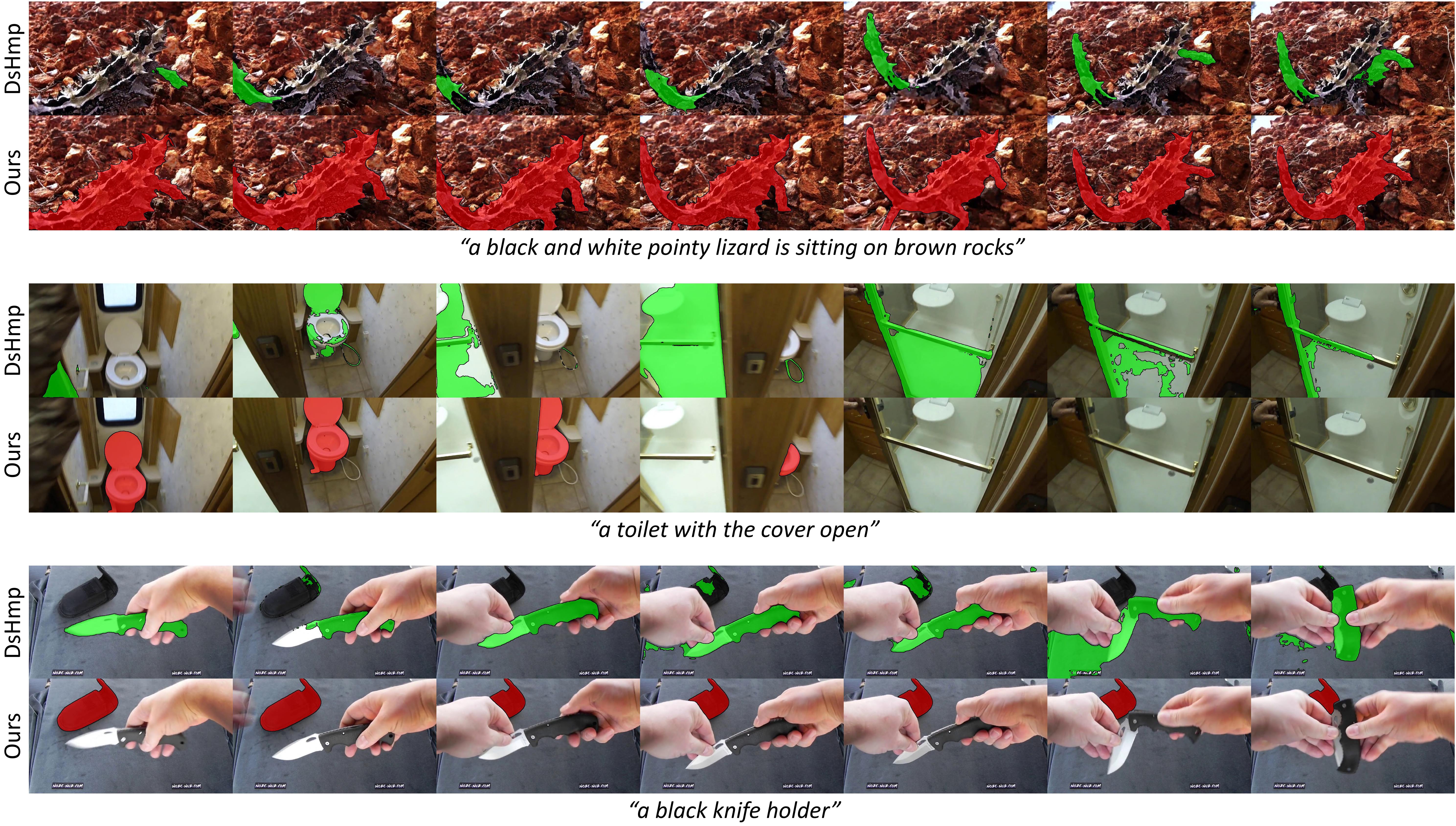}
    \vspace{-15pt}
    \caption{
    \textbf{Qualitative results on Ref-YouTube-VOS.} Our proposed method outperforms previous state-of-the-art approach~\cite{he2024decoupling} in terms of accurate selection of the corresponding object, while ensuring accurate segmentation of the corresponding object based on the given expression.}
    \label{fig:suppl_ytbvos_qual}
    \vspace{-10pt}
\end{figure*}

%% file: suppl/fig/noise_gaussian_qual.tex
\begin{figure*}[ht]
    \centering
    \vspace{-10pt}
    \includegraphics[width=\linewidth]{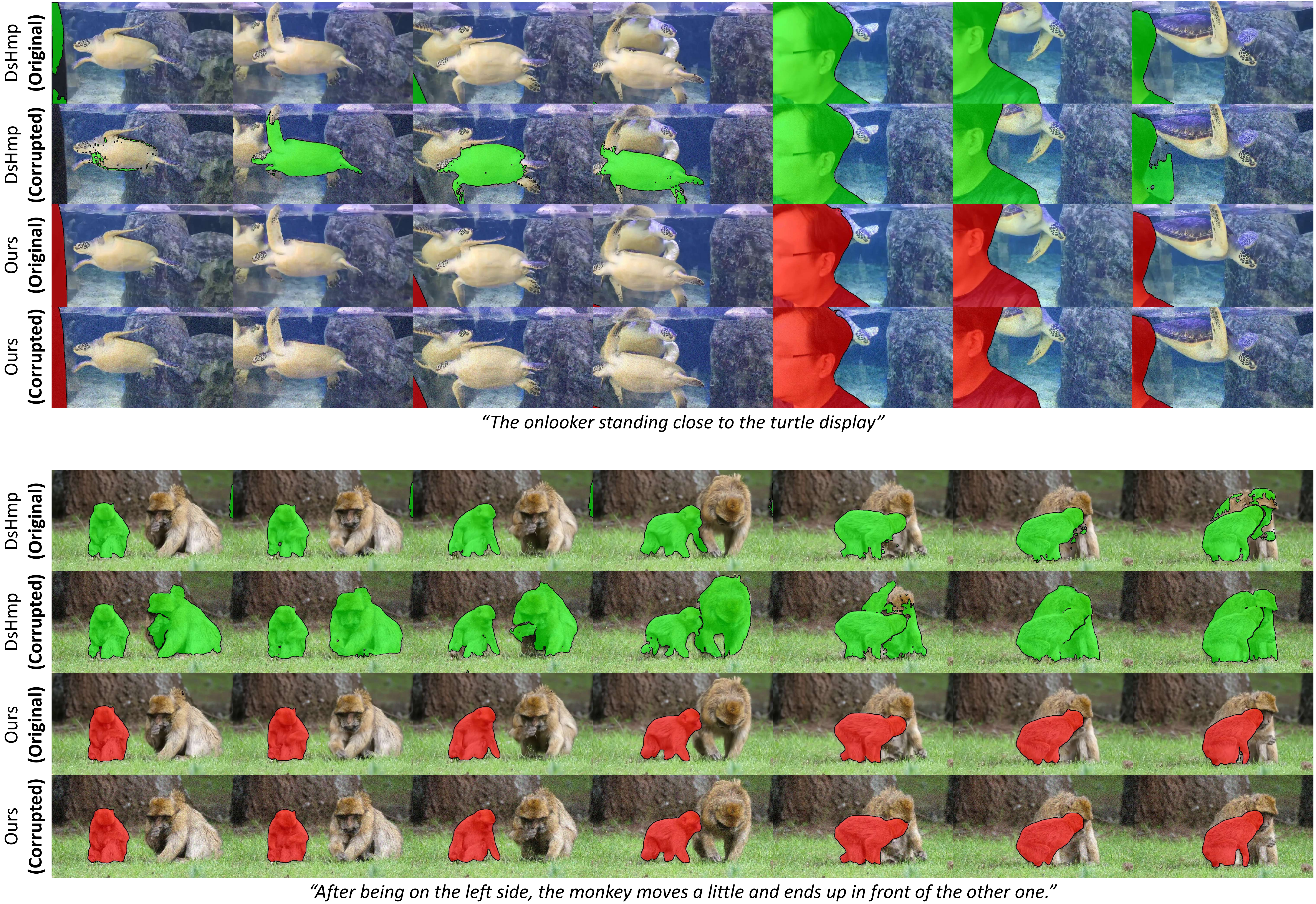}
    \vspace{-15pt}
\caption{\textbf{Qualitative results on corrupted version of MeViS.} Despite the \textit{gaussian noise} distortion, our method generates high-quality outputs, demonstrating its robustness and effectiveness in handling perturbed data. Compared to previous work, our results maintain their performance even under the corrupted setting.}
    \label{fig:suppl_gaussian_qual}
\end{figure*}

%% file: suppl/fig/noise_motion_qual.tex
\begin{figure*}[ht]
    \centering
    \vspace{-10pt}
    \includegraphics[width=\linewidth]{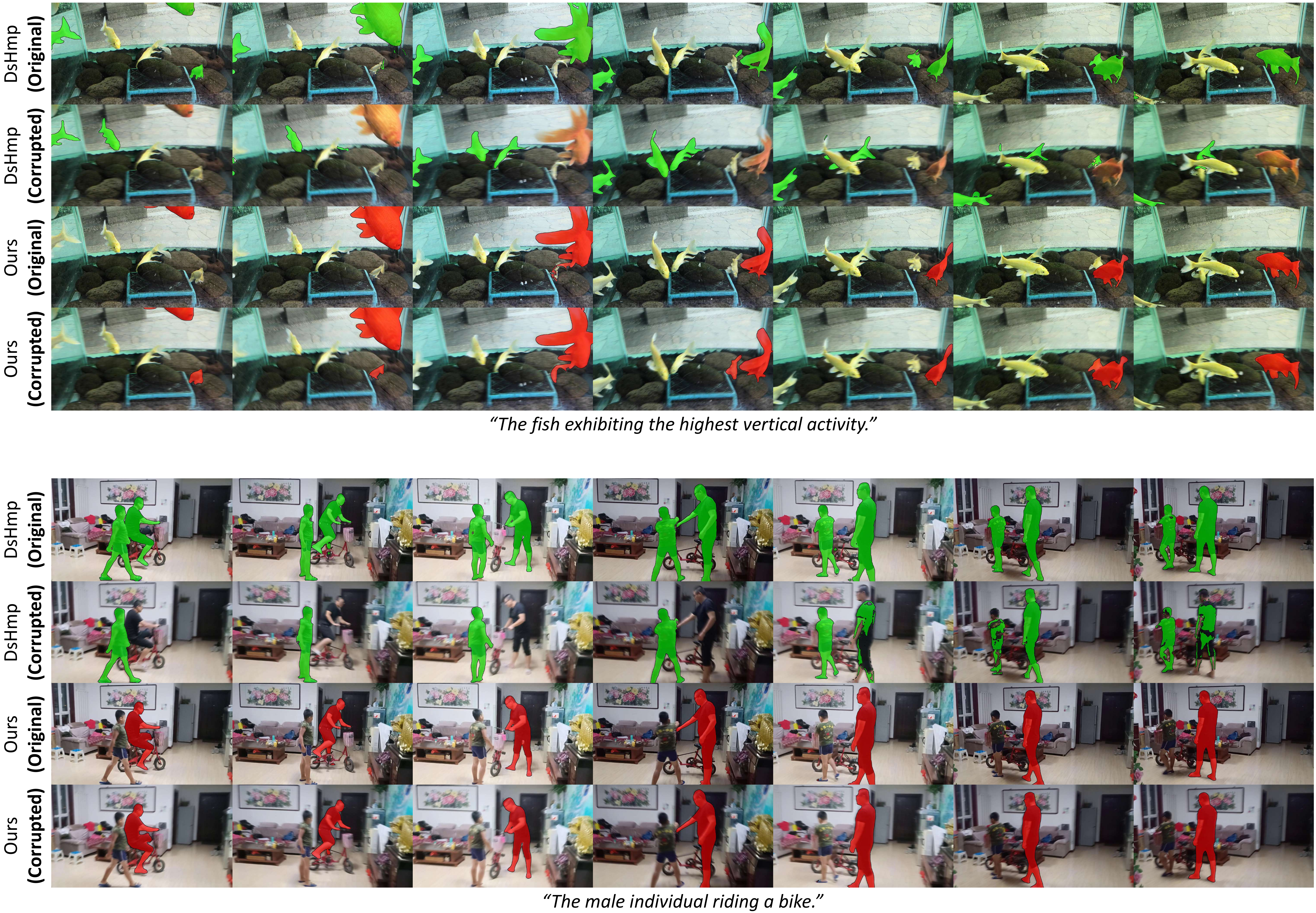}
    \vspace{-15pt}
\caption{\textbf{Qualitative results on corrupted version of MeViS.} Despite the \textit{motion blur} distortion, our method generates high-quality outputs, demonstrating its robustness and effectiveness in handling perturbed data. Compared to previous work, our results maintain their performance even under the corrupted setting.}
    \label{fig:suppl_motion_qual}
\end{figure*}